# Hybrid Improved Document-level Embedding (HIDE)

Satanik Mitra [1]*          Mamata Jenamani [2]

Research Scholar            Professor

satanikmitra@iitkgp.ac.in [1]    mj@iitkgp.ac.in[2]

Address: Department of Industrial & Systems Engineering, IIT Kharagpur,
Kharagpur - 721302, West Bengal, India

*Corresponding Author

**Abstract:** In recent times, word embeddings are taking a significant role in sentiment analysis. As the generation of word embeddings needs huge corpora, many applications use pre-trained embeddings. In spite of the success, word embeddings suffers from certain drawbacks such as – it does not capture sentiment information of a word, contextual information in terms of parts of speech tags and domain-specific information. In this work we propose HIDE a Hybrid Improved Document-level Embedding which incorporates *domain information*, *parts of speech information* and *sentiment information* into existing word embeddings such as GloVe and Word2Vec. It combine improved word embeddings into document level embeddings. Further, Latent Semantic Analysis (LSA) has been used to represent documents as a vectors. HIDE is generated, combining LSA and document level embeddings, which is computed from improved word embeddings. We test HIDE with six different datasets and shown considerable improvement over the accuracy of existing pre-trained word vectors such as - GloVe and Word2Vec. We further compare our work with two existing document-level sentiment analysis approaches. HIDE performs better than existing systems.

## 1. Introduction

Sentiment analysis is a technique to classify document based on latent opinion expressed in the textual contents. It used across a various domain of research, such as marketing decision making, social media analysis and many more (Deng, Sinha, & Zhao, 2017; Mostafa, 2013; Pantano, Giglio, & Dennis, 2019). With recent development in Natural Language Processing (NLP) techniques, word embedding is frequently used in many NLP tasks such as –syntactic parsing, question answering etc.(Iyyer, Boyd-Graber, Claudino, Socher, & Daumé, 2014;

Socher, Bauer, Manning, & Ng, 2013). Word embedding is a robust approach to capture semantic of a target word by finding conditional probability from its context (Shuang, Zhang, Loo, & Su, 2020). These embeddings represent words in a continuous low dimensional vector. As context of word is involved in generation of these vectors, the words with similar semantics are represented in near proximity in the embedding space. However, gerneration of word embedding require large corpora and involve costly processing time. This drawback compel researchers to turn their focus towards pre-trained word embeddings. Among pre-trained embeddings, Word2Vec and GloVe are used successfully(Corrêa & Amancio, 2019; Kameswara Sarma, 2018; Mikolov, Sutskever, Chen, Corrado, & Dean, 2013; Olatunji, Li, & Lam, 2020; Pennington, Socher, & Manning, 2014; Saumya, Singh, & Dwivedi, 2019).

In spite of the success of pre-trained embeddings in various NLP tasks, it is not considerably perform well in sentiment analysis. This performance degradation of pre-trained embeddings in sentiment analysis tasks is driven by certain shortcomings. Firstly, pre-trained embeddings are generated predominantly to capture the semantics of a word using its context. However, many times a positive and a negative word appear in similar word context. Secondly, words in a particular domain have a different distribution than the generic textual contents which is used to generate pre-trained embedding. Due to lack of domain-specific information, the performance of pre-trained embeddings in sentiment classification tasks is degrading (Blitzer, Dredze, & Pereira, 2007; Sarma, Liang, & Sethares, 2018). Thirdly, although word embeddings represent the semantics of a word based on its context, it does not incorporate the entire information concerning a word such as parts of speech (POS) (Petroni, Plachouras, Nugent, & Leidner, 2018).

In this work, we address the problems discussed above and present a Hybrid Improved Document-level Embedding (HIDE) by incorporating domain-specific, sentiment and POS information. Firstly, we adopt an approach to enrich word level embeddings with domain specific information. We consider pre-trained GloVe and Word2Vec embedding. Here, Word2Vec is generated from a domain-specific corpus. The combined vector of both word embeddings captures generic representation of words along with the domain information. Secondly, to address the word polarity related issue, we use lexicon-based sentiment analysis dictionaries to find the polarity of a word. We vectorized the lexicon information and embed it into word vectors. Thirdly, we integrate the parts of speech (POS) information of each word in a vectorized format. The document vector is generated by combining the embeddings of each word appeared in the document. Next, to incorporate document level domain information we use Latent Semantic Analysis (LSA) which represent the documents in low dimensional vector format. Combination of these two document level vector produces HIDE. We have tested our approach through several experiments with six benchmark datasets and Support Vector Machine (SVM) classifier with Gaussian radial basis kernel. The result shows an improvement over the pre-trained vectors as well as in sentiment analysis performance.

The work of Rezaeinia et al., (2019) motivates our approach of enriching word embeddings only; however, we differ in the following ways. First, we take combination of GloVe and Word2Vec embeddings and avoid random vector generation. This combination improves the word vectors with more relevant embedding than a random one. Second, all the word in a document are represented by their corresponding embeddings and document embedding is computed using average embeddings of participant words. Due to this, the position of a word in the document becomes irrelevant. Third, we have used less number of lexicon dictionaries to capture word sentiment scores; we discuss this in the following sections. Lastly, in contrast to word level embedding, we implement a domain-specific document level embeddings. Our main contribution in this work is as follows –

- We propose HIDE, an approach to develop a document level embedding with sentiment, POS and domain specific information.

- We improve word embedding by incorporating domain specific, POS and sentiment information using less number of lexicon databases.

- To incorporate document level domain information Latent Semantic Analysis (LSA) has been used. To best of our knowledge this is first time LSA is combined with improved word embeddings.

- HIDE supersedes many of the existing word embedding based sentiment classification approaches. We present our work with six different benchmark datasets and compare it with other approaches of word embedding improvement and domain-specific sentiment analysis.

The next portion of the paper is organised as follows – section 2 related works, section 3 methodology, section 4 experiment, section 5 results and discussion and section 6 conclusion respectively.

## 2. Related Works

The related literature concerning word and document level embedding is reported in the next sub-sections.

**2.1 Word representation using Embeddings**

In natural language processing, word representation is considered as one of the vital tasks (Yu, Wang, Lai, & Zhang, 2018). Word embedding represent a word as a distributed vector using its context in a large corpus with the help of a neural network. The earliest work, in this area, suggests, generation of embedding using neural network language model (NNLM) leveraging the context of each word (Bengio, Ducharme, Vincent, & Jauvin, 2003; Yu, et al., 2017). Later on, word embeddings are generated using models like a continuous bag of words (CBOW) and continuous skip-gram (Stein, Jaques, & Valiati, 2019). Both of these methods are concerned with the context of the word. In skip-gram, the current word is used to predict the context of the word and CBOW model, predict the word based upon its context (Pham & Le, 2018).

Word2Vec (Mikolov et al., 2013) and GloVe (Pennington et al., 2014) embeddings are used in many text classification tasks (Joulin, Grave, Bojanowski, & Mikolov, 2017; Rezaeinia et al., 2019; Yu, et al., 2017). Word2Vec is estimated by maximizing the log conditional probability, occurred within a window of context (Nalisnick, Mitra, Craswell, & Caruana, 2016). Whereas, GloVe works on global word co-occurrence and local context window (Pennington et al., 2014; Yu, et al., 2017). Generation of these custom embeddings requires large corpora, which is not always feasible, hence pre-trained Word2Vec and GloVe are used (Corrêa & Amancio, 2019; Kameswara Sarma, 2018; Mikolov et al., 2013; Pennington et al., 2014; Saumya et al., 2019). Pre-trained embedding computes the word vectors from general-purpose corpora such as Wikipedia (Rezaeinia et al., 2019). Although pre-trained embeddings are very efficient in text-related tasks, its performance is not appreciable in case of sentiment analysis (Kameswara Sarma, 2018; Rezaeinia et al., 2019; Sarma et al., 2018). This happens because these embeddings are created by prioritising the context of a word, and in that way, if two mutually opposite sentiment words occur in the same context then both of the words get similar embedding (Fu, Sun, Wu, Cui, & Huang, 2018; Tang et al., 2016). For example, words like *happy* and *sad* may appear in the same context (Yu et al., 2017). Unable to separate words based upon polarity degrades sentiment classification accuracy (Giatsoglou et al., 2017; Yu et al., 2017).

To address this problem, the sentiment embedding technique is proposed to refine the pre-trained word vectors using the lexicon sentiment intensity scores ( Yu et al., 2017). Giatsoglou et al., (2017) proposed a hybrid feature taking lexicon values along with word embeddings together. These works focus on incorporating sentiment information into word embeddings, but it does not archive parts of speeches (POS) information of the words. Att2vec method learn word embeddings, including the POS tag information (Petroni et al., 2018). The drawback here is it does not include lexicon information. In this context, IWV (Improved Word Vector) incorporates both POS tags and lexicon information into pre-trained word embeddings to improve the sentiment analysis accuracy (Rezaeinia et al., 2019).

Apart from the sentiment and POS related issues, pre-trained word embeddings also lack domain-specific information as they are generated out of general-purpose corpora (Dragoni & Petrucci, 2017; Sarma et al., 2018). To assimilate domain information word embeddings are to be computed from domain-specific corpora. Moreover, documents are built upon words; the existing approaches to combine word embeddings into the document is discussed in the next subsection.

## 2.2 Document-level Embedding

In supervised sentiment analysis techniques, term frequencies and count vectors are used as features in many of the cases (Tripathy, Agrawal, & Rath, 2016). These two methods have their own disadvantages. The vector space model generated from term frequency or by count vectors is high dimensional and heavily sparse. Moreover, it represents documents statistically rather than semantically and ordering of the words are not maintained (Huang, Qiu, & Huang, 2014; Stein et al., 2019). Word embeddings are useful in this context, but representing documents using word embedding is tricky as the frequency of words varies in different textual contents (Huang et al., 2014). To pacify this, an average of word vectors of each word present in the document is adopted by many researchers (Huang et al., 2014; Kameswara Sarma, 2018; Stein et al., 2019). Le & Mikolov, (2014) proposed paragraph vector, which is a unsupervised approach to representation fixed-length vector from sentences, paragraphs, and documents preserving word ordering. However, in paragraph vector the necessity of domain knowledge is not embedded. Domain knowledge plays a significant role in sentiment analysis. So, accumulation of domain knowledge and integrate into sentiment classification feature is important (Dragoni & Petrucci, 2017). Training of word vectors demands vast corpus. However, embedding with smaller dimension can capture a legitimate amount of information from a small-sized domain-specific corpus (Stein et al., 2019).

On the other hand, Latent Semantic Analysis (LSA) is used to produce a document level vector which is more semantically enriched way (Mitra & Jenamani, 2018; Stein et al., 2019). LSA transform the document term matrix in the denser low dimensional representation of the document using Singular Value Decomposition (SVD). When applied on documents of the same domain, LSA can capture domain-specific semantics in terms of vector representation (Fernández-Reyes & Shinde, 2019; Kameswara Sarma, 2018). It is shown that aligning pre-trained vectors along with domain-specific vectors generated using LSA, performs better compared to individual ones (Sarma et al., 2018). However, the lexicon-based sentiment scores of each word is not considered in this method. Moreover, word-level improvement of pre-trained embedding is not considered here. Table1 discussed key literature of this study. Table1 represents literature based on their proposed methods and used information to generate word embeddings.

**Table1: Few recent works to improve word embedding**

| Paper | Proposed Method | Used information | | |
|---|---|---|---|---|
| | | **Domain** | **Sentiment** | **POS** |
| (Mikolov et al., 2013) | Word2Vec model where word vectors represented by maximizing the log conditional probabilities within a context window. | X | X | X |
| (Pennington et al., 2014) | The GloVe uses the global co-occurrence matrix as well as local contexts. | X | X | X |
| (Giatsoglou et al., 2017) | Proposed a hybrid vector with vector produced by lexical scores and pre-trained word embedding. | X | ✓ | X |
| (Dragoni & Petrucci, 2017) | NeuroSent tool has been proposed for domain-specific polarity classification. A domain-specific corpus is used to generate word embeddings. | ✓ | X | X |
| (Yu et al., 2017) | Sentiment embeddings are generated using word embeddings and lexicon scores of words. | X | ✓ | X |
| (Petroni et al., 2018) | Used POS tag information to contextually enrich word embeddings for better classification performance. | X | X | ✓ |
| (Sarma et al., 2018) | Propose domain adapted word embedding where LSA based domain-specific embedding and pre-trained vectors are aligned using canonical correlation analysis. | ✓ | X | X |
| (Rezaeinia et al., 2019) | Focused on improving the pre-trained word vectors for better accuracy in sentiment analysis tasks. | X | ✓ | ✓ |
| Our Approach | LSA based domain-specific embeddings combined with improved pre-trained and domain-specific word embeddings to analyse the sentiment of a document | ✓ | ✓ | ✓ |

## 3. Methodology

In HIDE we improve the performance of word embeddings by applying natural language processing techniques, lexicon score of words, parts of speech information of the words. We configure document level embedding from these word embeddings and combined with domain-specific document level vectors to improve the sentiment classification tasks. In Fig.1, we show our proposed methodology. We improve word embeddings with domain-specific corpora and sentiment information. Then we generate document embedding from improved embeddings. We perform LSA to capture document level domain adaptive representation. Finally, we combined these two document embeddings into HIDE model.

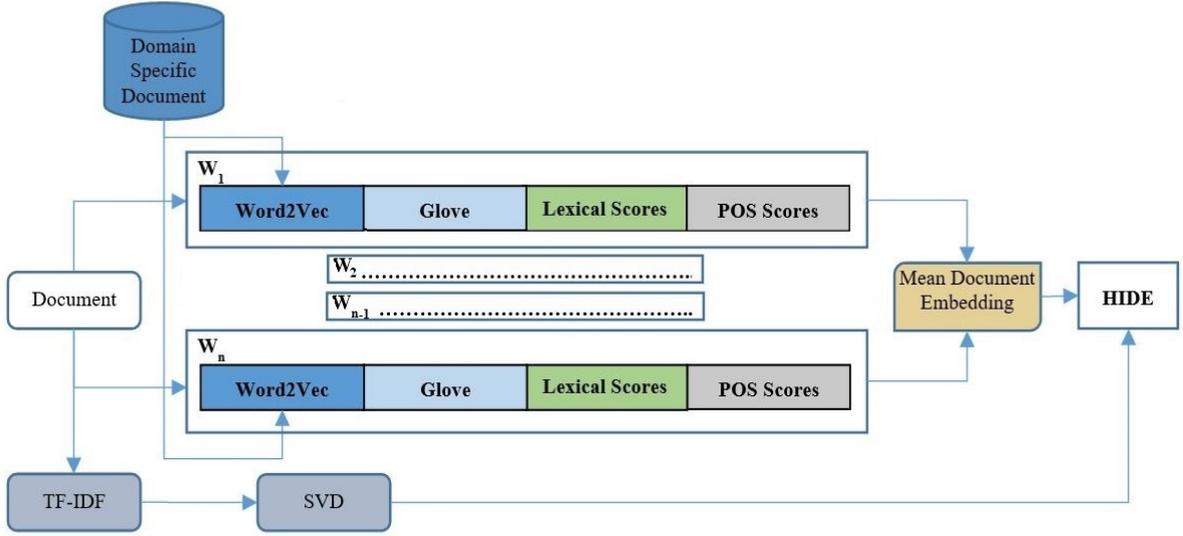

Fig.1: The main framework for Hybrid Improved Document-level Embeddings (HIDE)

### 3.1 Generation of word embedding

Pre-trained word embedding do not have domain-specific information as it is trained on large corpora of diverse topics, mostly on Wikipedia corpus. To address this issue we have used Word2Vec on domain-specific corpora. Use of domain specific texts reveals domain-specific semantics which embed into the word vectors (Ferrari, Donati, & Gnesi, 2017; Hu et al., 2018). Gensim python library has been used in this purpose. This library provides modules to generate Word2Vec and related algorithms. Now, In comparison with other general-purpose corpora, domain-specific corpora is not huge. Small dimensional word vectors works better to capture semantics, if small size dataset has been used for training. Hence, we generate 50-dimensional domain-specific embedding. We use separate domain specific corpora to generate domain-specific Word2Vec embeddings. The description of domain specific corpora is given in the next section. Along with domain specific Word2Vec, we incorporate 50 dimensional GloVe pre-trained embeddings.

Every word in the classification process is converted into its corresponding embedding by taking average of domain-specific Word2Vec and pre-trained GloVe. However, there may exist some of new words in the classification dataset which do not appear in the domain specific corpora. To represent those new words only GloVe pre-trained embedding has been used. Here, say $w$ is a word in the sentiment classification corpus. Word embeddings coming from domain-specific corpora using Word2Vec method are represented by $w_{word2vec}$. The notation $w_{glove}$ represents the corresponding pre-trained GloVe embedding. To get the GloVe pre-train embedding for $w$, $w_{glove}(w)$ method is used. Similarly, $w_{word2vec}(w)$ returns the domain-specific Word2Vec embedding of $w$. The combined embedding for $w$ is stored in $w_{combine\_embedding}(w)$. Eq.1 explains the $w_{combine\_embedding}(w)$ generation process.

$$w_{Word2Vec}(w) = w_{Word2Vec}(domain\ corpora)$$
$$w_{combine\_embedding}(w) = avg(w_{GloVe}(w), w_{Word2Vec}(w)),$$
$$\text{where } w_{GloVe}(w), w_{Word2Vec}(w) \neq \Phi$$
$$w_{combine\_embedding}(w) = w_{GloVe}(w),\ \text{where } w_{Word2Vec}(w) = \Phi \quad (1)$$
$$w_{combine\_embedding}(w) = w_{Word2Vec}(w),\ \text{where } w_{GloVe}(w) = \Phi$$

For example, suppose a sentence present in the corpus as *"It's a great movie to watch"*. So, after removing the stop words, i.e. *"it's", "a", "to"* there remains three words *"great", "movie"* and *"watch"* which can polarize the sentence. Table2 shows the method of combining Word2Vec and Glove.

**Table2: Example of combining word embedding**

| Words | Great | Movie | Watch |
|---|---|---|---|
| Word2Vec | $w_{Word2Vec}(Great)$ | $w_{Word2Vec}(Movie)$ | $w_{Word2Vec}(Watch)$ |
| GloVe | $w_{GloVe}(Great)$ | $w_{GloVe}(Movie)$ | $w_{GloVe}(Watch)$ |
| Combined Embedding | $w_{combine\_embedding}(Great)$ | $w_{combine\_embedding}(Movie)$ | $w_{combine\_embedding}(Watch)$ |

## 3.2 Improving word embedding

The improvement of word embedding is done using sentiment lexicon scores; POS tag information.

### 3.2.1 Lexical Score to Vector (Lex2vec)

Sentiment lexicons are the words which are vital for predicting the sentiment of a document (Medhat, Hassan, & Korashy, 2014; Turney, 2001). The lexicon-based approach identifies the polarity of a document using the semantics of the phrases present (Taboada, Brooke, Tofiloski, Voll, & Stede, 2011). Here a dictionary of sentiment words is used along with the score for each lexicon. Many lexicon dictionaries are available. From them, choosing a lexicon dictionary is vital as because some of the dictionaries are more accurate than the other.

1. SemEval-2015 English Twitter Sentiment Lexicon (Kiritchenko, Zhu, & Mohammad, 2014; Rosenthal et al., 2015)
2. National Research Council Canada (NRC) –Amazon laptops Sentiment Lexicons for Unigrams (Kiritchenko, Zhu, Cherry, & Mohammad, 2015)

SemEval-2015 English Twitter Sentiment Lexicon (Kiritchenko, Zhu, & Mohammad, 2014; Rosenthal et al., 2015) contains 1515 lexicons and value ranges from -0.984 to +0.984 which capable of identifying smaller changes in sentiment. The second lexicon database is National Research Council Canada (NRC) –Amazon laptops Sentiment Lexicons for Unigrams (Kiritchenko et al., 2015) which contains a total of 26,577 lexicons and value ranges from -5.27 to +3.702. Each of these lexicon databases has a word or a phrase along with its sentiment scores. We normalized and scaled the sentiment scores of all the words between [0, 1] and formed a vector by concatenating the lexical values from both these databases. The lexical values of the words not present in the databases are considered to be zero. The lexical score vector of running example sentence is given in Table3.

**Table3: Running Example of Lexical Score to Vector Method**

| Words | NRC)Amazon laptops Sentiment Lexicons | SemEval-2015 English Twitter Sentiment Lexicon | Lexical Score Vector | Normalized Lexical Score Vector $\{w_{lex}\}$ |
|---|---|---|---|---|
| *great* | +0.957 | +0.734 | [0.957, 0.734] | [0.9785, 0.867] |
| *movie* | +0.610 | +0.188 | [0.610, 0.188] | [0.805, 0.594] |
| *watch* | +0.809 | +0.266 | [0.809, 0.266] | [0.9045, 0.633] |

### 3.2.2 Parts Of Speech to Vector

POS tagging is a technique of assigning the words with their Part of Speech tags. This information is not captured in the pre-trained word embedding. However, the recent study shows it improves the classification accuracy of pre-trained embedding (Petroni et al., 2018). In case of sentiment analysis, involving POS tag information in feature vector gives better results (Pasupa & Seneewong Na Ayutthaya, 2019). The POS tag helps the model to figure out the actual meaning of the word (whether the word is a noun, adverb, adjective etc.). We figure out the POS tags of a word using a POS tagger and converted the POS tag of each word into a vector having one-hot representation, i.e. the index of the corresponding tag was given a value one and all others as zero. Nltk toolkit (Navarre & Steiman, 2002) considers 36 different POS tags and so we used 36 different values in $w_{pos}$ the vector. Eq.2 shows the derivation of $w_{pos}$.

$$w_{pos} = [<pos\_tag_1>, <pos\_tag_2>, ....., <pos\_tag_n>] \quad (2)$$

where $n = |pos\_tag|$

$pos\_tag_i = 1$ if $pos\_tag_i = pos\_tag(w_i)$

$pos\_tag_i = 0$ where $pos\_tag_i \neq pos\_tag(w_i)$

So, for the running example discussed earlier the corresponding parts of speech to vector are described in Table4.

**Table4: Example of Parts of Speech to Vector**

| Words | Great | Movie | Watch |
|---|---|---|---|
| POS Tags | Adjective*(JJ)* | Noun*(NN)* | Verb*(VB)* |
| $\{w_{pos}\}$ | $\{w_{JJ}\}$ | $\{w_{NN}\}$ | $\{w_{VB}\}$ |

## 3.3 Final Embedding of Each Word

To get the final improved embeddings of each word $w_{final\_embedding}$, we combined lexical vectors $\{w_{lex}\}$, POS vectors $\{w_{pos}\}$ and embedding generated by combination of Word2Vec and GloVe embeddings $w_{combine\_embedding}$. Fig.2 shows the final embedding of the running example. Fig.3 displays the entire process of improving word embeddings. As both Word2Vec and GloVe pre-train embeddings are of same length of 50-dimesion the resultant average combination is of same size. However, after appending the lexical and POS vector the final embedding increase in length. We rescale the dimensions of embeddings in Fig.3 for the ease of understanding. These improved embeddings are used further to generate document level embeddings.

$$w_{final\_embedding}(Great) = \{w_{combine\_embedding}(Great), w_{lex}(Great), w_{pos}(Great)\}$$

$$w_{final\_embedding}(Movie) = \{w_{combine\_embedding}(Movie), w_{lex}(Movie), w_{pos}(Movie)\}$$

$$w_{final\_embedding}(Watch) = \{w_{combine\_embedding}(Watch), w_{lex}(Watch), w_{pos}(Watch)\}$$

Fig.2 final embedding of the running example

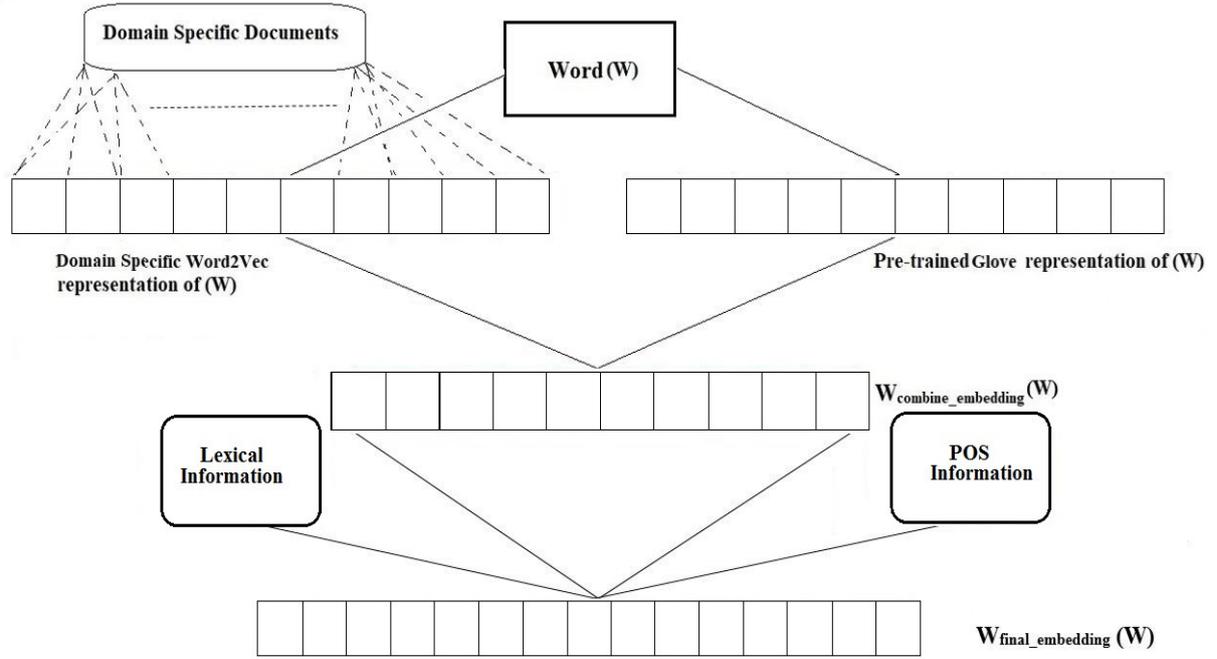

Fig.3 Improved embedding of each words

### 3.4 Document Representation using Word Embedding

Documents are collection of words. With word embedding technique we get embeddings of each word. As we want to incorporate the domain information, we generate our document embedding by combining two types of vector representation. First, document embedding is generated from the improved word vectors. Second, another document representation is computed by applying LSA on the training corpus. The final document embedding is generated by combining these two document representation. GloVe and Word2Vec do not provide any optimized embedding for sentences or documents, so taking the average of embeddings corresponds to all words present in a document can give a document level embedding (Renter, Borisov, & De Rijke, 2016). The document embedding is shown using $doc_{embedding}$. Eq.3 display the computation of $doc_{embedding}$, where $w_{final\_embedding(i)}$ shows the embeddings of each words present in a document and $N$ is the total number of words present in the document.

$$doc_{embedding} = \frac{\sum_{i=1}^{N}\left(w_{final\_embedding(i)}\right)}{N} \qquad (3)$$

On the other hand, as mentioned in section 2.2 LSA is a way to keep the domain information in the document embedding. Here, LSA of the document $doc_{LSA}$ is generated by taking the Singular Value Decomposition (SVD) of the Term Frequency-Inverse Document Frequency (TF-IDF) matrix. This document vector is smaller in dimension and denser than TF-IDF vectors.

Now, HIDE is generated by combining $doc_{embedding}$, $doc_{LSA}$. We combine these two vectors. We linearly append these vectors to form HIDE. Eq.4 shows the process. The HIDE generation algorithm is shown in Algorithm1.

$$HIDE = \{doc_{embedding}, doc_{LSA}\} \quad (4)$$

## 4. Experiments

This section is about the experimentation conducted to analyse the performance of HIDE on six benchmark datasets. The description of the datasets, our approach and setting of our experimentation has been discussed. We primarily estimate the advantage of HIDE over the use of word vector and lexicons. Different combination of word embedding and lexicon database has been used to generate hybrid document level vectors. We also compare HIDE with existing method to understand its efficiency.

### 4.1 Description of Datasets

We use total of six benchmark datasets. We take datasets from disjoint domains such as – movies, electronic products etc. The datasets are –

**MR:** This dataset contains 1000 positive and 1000 negative movie reviews categorised based on their star ratings (Pang & Lee, 2005). Rating less than 3, labelled as negative and rating four and five marked as positive.

**CR:** It contains reviews of 14 different products and labelled as positive and negative(M. Hu & Liu, 2004).

**RT:** It is having 5331 positive and same number of negative reviews which introduced by Pang & Lee, (2005).

**SST**: Stanford Sentiment Treebank dataset consisting 11,855 sentences of label positive and negative (Socher, Perelygin, et al., 2013).

**SST-1:** This dataset is similar to SST having train, test and dev splits and includes fine-grained labels such as - very positive, positive, neutral, negative, very negative (Socher et al., 2013).

**Amazon Review Dataset (ARD):** This dataset contain 14.2 million entries (He & McAuley, 2016; McAuley, Targett, Shi, & Van Den Hengel, 2015). Although, the reviews are not labelled as per positive or negative sentiment but in the light of Pang & Lee, (2005) we labelled these reviews in positive and negative categories based on their star rating. Rating below three considered as negative and above three marks as positive.

The description of dataset has been given in Table5. We use benchmark datasets from the movie and electronic domain to generate Word2Vec embeddings. In this purpose, we use Large Movie Review dataset for binary sentiment classification (Maas et al., 2011). It contains 25,000 training and 25,000 for testing reviews. For electronic domain Amazon review dataset (He & McAuley, 2016; McAuley et al., 2015) has been used. The electronic category contains around 1,689,188 reviews. We randomly select 25000 reviews for domain specific Word2Vec generation. The detail description of datasets used in both experiments and domain specific Word2Vec generation is given in Table6.

**Table5: Description of the datasets used in experiment**

| Dataset Name | Average Words per Sentence | Number of Reviews | Unique Words | Sentiment Classes |
|---|---|---|---|---|
| MR | 12 | 2000 | 26675 | 2 |
| CR | 19 | 3775 | 5340 | 2 |
| RT | 20 | 10662 | 18765 | 2 |
| SST 1 | 18 | 11855 | 17836 | 5 |
| SST | 19 | 9613 | 16185 | 2 |
| ARD | 11 | 5000 | 32721 | 2 |

**Table6: Description of the domain specific dataset**

| Dataset Name | Avg Words per Sentence | Number of Reviews | Unique Words | Avg Review Length |
|---|---|---|---|---|
| Amazon | 11 | 10000 | 40931 | 86 |
| IMDB | 3 | 10000 | 49255 | 123 |

## 4.2 Evaluation protocol

We present HIDE with Support Vector Machine (SVM) classifier. It is a discriminative classifier which realized by a separating hyper-plane. In many of text classification tasks along with sentiment analysis SVM has been used (Catal & Nangir, 2017; Dey, Jenamani, & Thakkar, 2017; Tripathy et al., 2016). Accuracy is used to assess the performance of the method which is defined as –

$$Accuracy = \frac{TP + TN}{TP + TN + FP + FN} \quad (5)$$

Where, $TP$ is True positive, which represent reviews labelled correctly, as positive. True Negative represents by $TN$ shows negative reviews labelled with negative. The other conditions are represented by False Positive ($FP$) and False Negative ($FN$) when negative reviews are labelled as positive and positive reviews are labelled as negative respectively. Let's consider, two sentences from MR dataset, "*I recommend this movie for children and adults who are a child at heart*" and "*The film had no script to test any actors acting skill or ability*". It is clear that the first sentence is belong to the positive class and second one belongs to the negative class. Now, if the model classifies the first sentence as positive, it will be consider as True Positive (TP) and if the second sentence is classified as negative then it will be True Negative (TN). However, False Positive (FP) happens when the second sentence is labelled a positive. It is also similar to type I error. Similarly, if the first sentence marked as negative by the model it will be a False Negative (FN) incident, which may refer as type II error as well. We can represent the classification scenario using a 2x2 confusion matrix. Fig.4 shows the corresponding confusion matrix. We use Precision (P), Recall (R) and F-scores measures based on the confusion matrix to get more precise results. Precision, recall and F-score defined as –

$$Precision = \frac{TP}{TP + FP} \quad (6)$$

$$Recall = \frac{TP}{TP + FN} \quad (7)$$

$$F\text{-}score = \frac{2 \times (P \times R)}{P + R} \quad (8)$$

|  | Predicted | |
|---|---|---|
|  | **Positive** | **Negative** |
| **Actual Positive** | *I recommend this movie for children and adults who are a child at heart* (**TP - True Positive**) | *I recommend this movie for children and adults who are a child at heart* (**FN -False Negative**) |
| **Actual Negative** | *The film had no script to test any actors acting skill or ability* (**FP - False Positive**) | *The film had no script to test any actors acting skill or ability* (**TN - True Negative**) |

Fig.4: Confusion Matrix

---

**Algorithm1: HIDE generation technique**

**Input:**
$D_{ds}$ = { $Doc_1, Doc_2, Doc_3 \ldots Doc_m$ } , *Domain specific corpora*
$D_I$ = { $Doc_1, Doc_2, Doc_3 \ldots Doc_n$ }, *Input Datasets*
| $D_{ds}$ | = m, |$D_I$| = n
$D_{ds}\_W2V$ = Word2Vec trained on $D_{ds}$ Domain specific corpora
Glv = Pre-trained GloVe embeddings
PT = { $Tag_1, Tag_2, Tag_3 \ldots Tag_k$ }, *All Parts of Speech Tags*
Set of lexicon library = { $lex_1, lex_2$ }, *All lexicons*

**Output:**
HIDE: Hybrid Improved Document Level Embeddings

**Begin:**
  $Doc_{embedding}$ = [ ] empty list to keep document embeddings
  *for* **each** *d* in $D_I$:
    A = [ ]                              //*empty list to store embeddings of each word in each document*
    *for each* word *w* in *d*:
      *lex2vec* =[ ]                  //*empty list to keep lexicon vectors*
      *pos_tag* = [ ]                //*empty list to keep POS vectors*
      $W_{combine\_embedding}$ = [ ]       //*empty list to keep vectors from Word2Vec and Glove combination*
      *if* *w* exists in $D_{ds}\_W2V$ and Glv:
        $W_{combine\_embedding}(w).append((D_{ds}\_W2V(w) + Glv(w))/2)$
      *elif* *w* exists in $D_{ds}\_W2V$ only:
        $W_{combine\_embedding}(w).append(D_{ds}\_W2V(w))$
      *elif* *w* exists in Glv only:
        $W_{combine\_embedding}(w).append(Glv(w))$
      *end if*
    $POS\_tag$ = $pos\_tag_i(w)$)
    *for pos = 1 to k:*                // *k is the total number of POS tag used*
      *if pos= $Tag_k$ in POS_tag:*
        $pos\_tag.append(V_{Tagk})$
      *end if*
    *end for*
    *for lex=1 to 2:*
      *if w in $lex_i$:*
      *lex2vec.append* ($Normalized(lex_i(w))$
      *end if*
    *end for*

```
    A.append(W_{combine\_embedding} (w), pos\_tag, lex2vec)
  end for
    Doc_{embedding}.append((A))
    D_{tf-idf} = Tf-Idf (D_I)                    // Term frequency Inverse Document Frequency generation
    D_{LSA} = SVD(D_{tf-idf})                    // Single Value Decomposition for Latent Semantic Analysis(LSA)
    HIDE = {Doc_{embedding} , D_{LSA}}           //concatenation of Doc_{embedding} and D_{LSA}
    Return HIDE
```

## 5. Results and Discussion

In next subsections, we report the results obtained by applying the evaluation protocol stated above on six datasets. We experiment with three popular machine learning classifiers and choose SVM. As HIDE is a document level representation and the basic intention is to estimate its performance over the pre-trained vectors, we experiment with document vector generated using only GloVe and domain specific Word2Vec separately. Moreover, HIDE considers two types of lexicons, hence, we experiment with individual lexicon databases along with their combinations to see how performance varies. We compare HIDE with two existing models. First, domain adapted sentiment analysis model proposed by Kameswara Sarma, (2018). Second, with improved word vectors (IWV) (Rezaeinia et al., 2019). Table8 shows the performance scores. Cross validation is applied whenever is possible. SST and SST-1 datasets are already clubbed into train and test set, so we do not use further cross validation. However, to find best classification parameter we perform 10-fold cross validation. To find suitable parameter for SVM we use grid search algorithm with parameter C and gamma. Best possible performance has been chosen based on performance accuracy in the grid search.

We have used python as the basic programming language for experimentation. We have used 50 dimensional GloVe embeddings. To keep the consistency in vector length the dimension of the Word2Vec is remain 50. To represent the POS information 36 dimensional vectors has been used. Further for lexicon information for each word a 2 dimensional vector is computed. Lastly, document representation using LSA is computed into a 50 dimensional space to keep uniformity. Finally, after combining all these vectors we get a 138 dimensional HIDE embedding. These vectors are considered as final feature vector and use it for classification.

### 5.1 Experiment with other classifiers

SVM is used as base classifier in HIDE. To understand the linearity of the final HIDE embedding we implement t-SNE plots with randomly selected 70 documents from positive and negative data. TSNE, t-Distributed Stochastic Neighbour Embedding, is a technique to visualize high dimensional data. Fig.5 shows the t-SNE plot of HIDE for four datasets. The t-SNE plot indicates that the HIDE is not linearly separable. Hence, we use Gaussian radial basis function as SVM kernel instead of linear one. However, before choosing SVM as classifier, we conduct experiments with other popular machine learning classifiers like – Naïve Bayes, Decision Tree. These classifiers are experimented upon MR and RT datasets. Neural networks and deep learning models are not used in our approach mainly because of two reasons. Firstly, neural networks works well with large amount of data. HIDE is a document level representation and as the dataset use in the process are small in size, we drop the implementation of neural networks. Secondly, neural networks are costly in terms of time and resource utilization. The evaluation of SVM with other two classifiers shown in Fig.6. Table7 shows the results of different classifiers. We use SVM parameter gamma is kept in "scale" and C=1 for this experiment. It is evident that the accuracy of SVM is higher than rest of the two classifiers with a steady precision recall value. Hence, we choose SVM as classifier for this study.

**Table7: Performance analysis of classifiers using HIDE (%)**

| Classifiers | Precision | | Recall | | Accuracy | |
|---|---|---|---|---|---|---|
| | MR | RT | MR | RT | MR | RT |
| Naïve Bayes | 93 | 59 | 71 | 92 | 71 | 65 |
| Decision Tree | 66 | 71 | 65 | 76 | 66 | 73 |
| SVM | 87.52 | 83 | 85.60 | 86 | **87.52** | **83.50** |

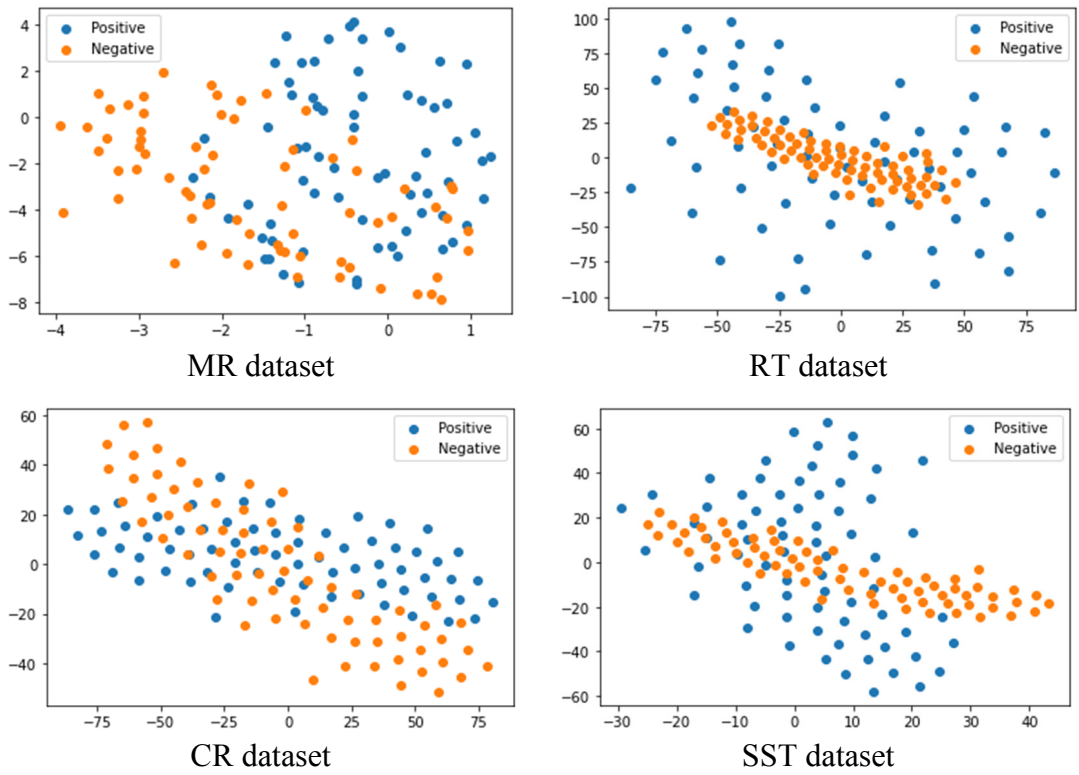

MR dataset

RT dataset

CR dataset

SST dataset

Fig.5: t-SNE plot of HIDE for four datasets

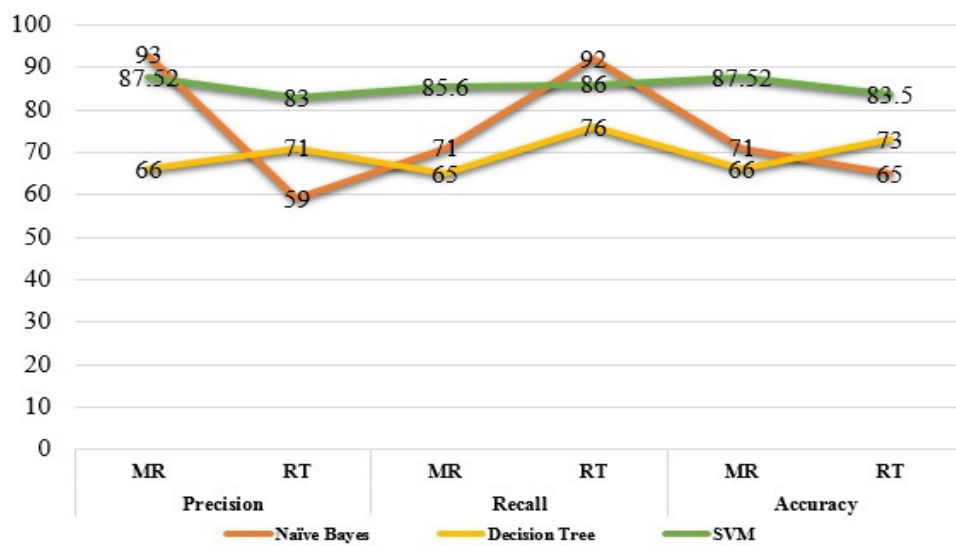

Fig.6: Comparison of SVM with Naïve Bayes and Decision Tree Classifiers

## 5.2 Experiment with Word Vectors

We experiment with three type of approaches to generate a document level embedding in context of HIDE. The first case involve only GloVe pre-trained vectors to generate document level embedding. The second case involve only Word2Vec, where it is trained on domain specific corpora. Lastly, we consider combination of both the embedding and generate document level embedding. Table8 shows the comparison based on accuracy. The precision, recall estimations are shown in Fig.7 and Fig8. Table9 represents the comparison based on f-scores. As per the grid search results we fix the SVM parameter for these experiment. For ARD the C value is 0.1 and for SST-1 gamma is 0.9 except that all the experiments give best results with C=1 and gamma= "scale".

Table8 clearly shows HIDE performs better than document embedding generated from GloVe, Word2Vec and their combination. It is also noticeable that combination of domain specific Word2Vec and GloVe does not provide better result than GloVe. It is because of two reasons. Firstly, Word2Vec is trained on domain specific corpora in comparison to the general purpose corpora, on which GloVe has been trained. As a result, Word2Vec captures essential domain specific information from the context around the target word. However, it significantly change the embedding of a particular word with respect to pre-trained GloVe. Secondly, due to the small size of domain specific corpora, it is not always possible to get embeddings of all the words present in the test dataset. This drawback is eminent in the performance of domain specific Word2Vec where its accuracy is way less than GloVe. HIDE eradicates this flaw by taking advantage of both the embeddings in association with lexical and POS information. Furthermore, HIDE combines document representation generated using LSA which contributes to its performance and makes it more befitting for sentiment analysis. HIDE performs maximum with IMDB (MR) dataset and Amazon Review Dataset (ARD) with 87% accuracy.

MR and RT are balanced datasets whereas CR is not a balanced dataset. GloVe performs better in CR with respect to other datasets. However, HIDE outperform GloVe with an increase of 3.5% in accuracy. Another interesting observation can be made from SST-1 dataset. This dataset contain multiple classes. As HIDE is modelled from the perspective of binary classification, it is expected it will perform with less accuracy with multiclass classification. Other vectors also suffers from similar drawback. The accuracy score of SST-1 for GloVe, Word2Vec and their combination remain close to HIDE. However, HIDE marginally performs better. F-scores for HIDE is better than GloVe and domain specific Word2Vec embeddings. Fig.5 shows the comparison of precision scores for different datasets. HIDE performs better than other method in terms of precision. Fig.6 shows the recall values for each dataset.

Table8: Comparison with pre-trained and domain specific word embeddings based on performance accuracy (%)

| Dataset | GloVe Pre-trained word vectors | Domain Specific Word2Vec | Domain Specific Word2Vec+GloVe | HIDE |
|---|---|---|---|---|
| MR | 79.00 | 59.00 | 52.50 | **87.52** |
| ARD | 78.42 | 53.57 | 70.00 | **88.14** |
| RT | 82.90 | 57.84 | 54.21 | **83.00** |
| CR | 77.50 | 70.67 | 60.00 | **81.67** |
| SST | 75.60 | 56.34 | 57.50 | **81.74** |
| SST-1 | 50.00 | 50.10 | 21.00 | **50.25** |

**Table9: Comparison based on f-score (%)**

| Dataset | GloVe Pre-trained word vectors | Domain Specific Word2Vec | Domain Specific Word2Vec+GloVe | HIDE |
|---|---|---|---|---|
| **MR** | 78.57 | 59.41 | 35.37 | **87.18** |
| **ARD** | 74.65 | 46.34 | 70.00 | **86.82** |
| **RT** | 81.13 | 66.68 | 68.06 | **83.50** |
| **CR** | 77.50 | 71.24 | 60.00 | **81.00** |
| **SST** | 71.33 | 47.22 | 62.53 | **79.32** |
| **SST-1** | 16.60 | 17.60 | 13.00 | **19.00** |

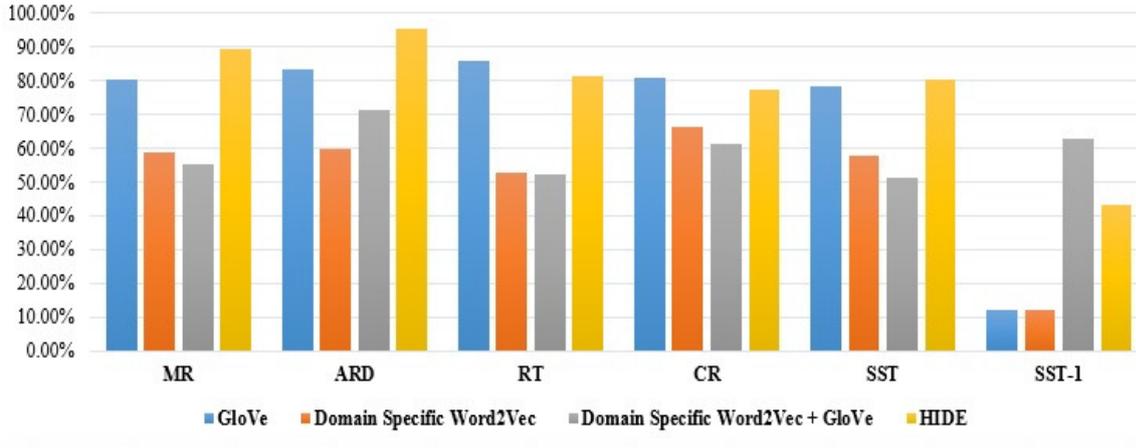

Fig.7: Comparison of HIDE with pre-trained and domain-specific word embeddings based on precision (%)

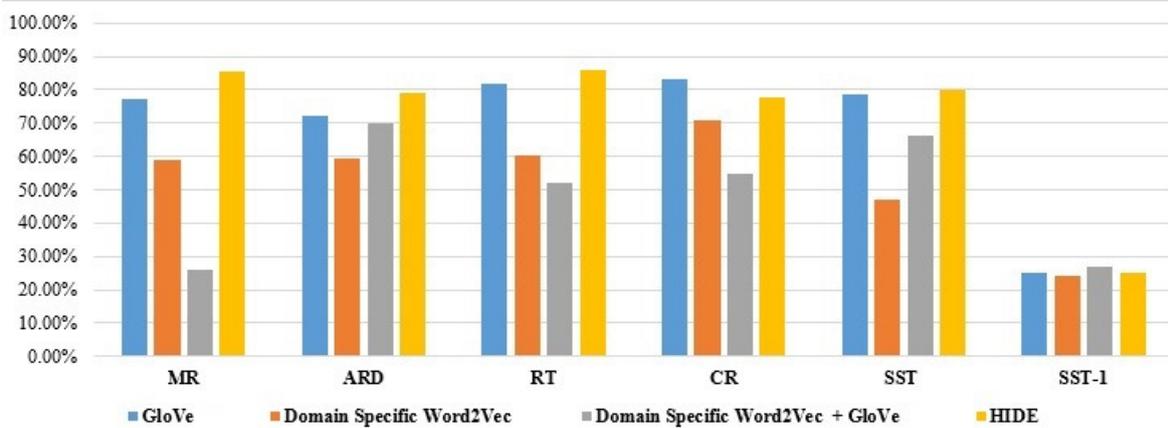

Fig.8: Comparison of HIDE with pre-trained and domain-specific word embeddings based on recall (%)

### 5.3 Experiment with Lexicon databases

The goal of this experiment is to analyse the effect of lexicon databases use to generate HIDE. The lexicon databases described earlier are marked as lex1 (SemEval-2015 English Twitter Sentiment Lexicon) and lex2 (NRC) respectively. To understand the contribution of each lexicon databases, we configure HIDE with individual lexical databases. In Table10 HIDE configure only with lex1 or lex2 is termed as HIDE+lex1 and HIDE+lex2 respectively. Experimenting with a lexicon database we keep the other lexical value as zero. An observation can be drawn from the scores of Table10, lex2 performs better in at least three datasets where the vocabulary size ranges between 20000. However, for datasets with larger vocabulary, lex1 performs better. Comparing the scores it is clear that, the best alternative is to consider both of the lexicons for better classification accuracy.

Table10: Comparison with two different lexicon datasets on performance accuracy (%)

| Dataset | HIDE+lex1 | | HIDE+lex2 | | HIDE | |
| --- | --- | --- | --- | --- | --- | --- |
| | Accuracy | F-Score | Accuracy | F-Score | Accuracy | F-Score |
| **MR** | 86 | 86 | 85 | 84 | **87.52** | **87.18** |
| **ARD** | 84 | 84 | 83 | 83 | **88.14** | **86.82** |
| **RT** | 73 | 72 | 76 | 76 | **83.00** | **83.50** |
| **CR** | 55 | 27.9 | 73 | 68 | **81.67** | **81.00** |
| **SST** | 59 | 63 | 73 | 74 | **81.74** | **79.32** |
| **SST-1** | 50 | 18 | 50 | 18 | **50.25** | **19.00** |

## 5.4 Comparison with other Methods

To analyse the performance of HIDE, we compare HIDE with existing word embedding improvement algorithm IWV (Rezaeinia et al., 2019) and Domain Adapted (DA) embeddings (Sarma et al., 2018). In IWV, pre-trained vectors, GloVe and Word2Vec has been improved with lexical information, POS tag and word position information. In contrast to IWV, in HIDE, document is represented by the average of word embeddings which makes word position information redundant. Moreover, domain specific Word2Vec is not implemented in IWV. In DA (Sarma et al., 2018) method, document embedding has been generated by LSA, then it combined with GloVe and Word2Vec pre-trained embedding to get higher accuracy in sentiment analysis. However, DA does not incorporates any lexical information.

In Table11 we show the comparitive scores between HIDE, IWV and DA. We compare HIDE with the scores reported in IWV for MR, CR, SST, SST-1, RT. Whereas, for ARD dataset we implement IWV method for comparison. To compare HIDE with DA method, we implement DA method for CR, SST, SST-1, RT, ARD datasets. It is evident from Table10, HIDE performs better than IWV and DA for both MR and RT datasets. HIDE improves approximately 7% over IWV and approximately 9% over DA method. Same thing is reflected in f-scores as well. For MR and RT dataset the HIDE shows improvement over IWV and DA for more than 5% and 7% respectively. In constrast to IWV and DA method, HIDE focused more on domain specific information by generating Word2Vec from domain specific corpora. Furthermore, LSA of domain specific training data is combined with HIDE. MR and RT bigger datasets in comparison with other datasets. It is quite evident that it contains more number of doamin specific words. For SST although IWV works better but HIDE outperforms DA by a difference of 20% approximately. In CR dataset HIDE again overruled DA with large margin but IWV performs better than HIDE. For both SST and CR the dataset is represented in sentential format, where each sentence is marked as positive and negative. So, with HIDE when we are applying LSA considering each sentences as a document the generated document embedding is not able to capture much information at document level. This is the cause of the visible drop in the accuracy of HIDE in both CR and SST dataset. In case of SST-1 dataset the accuracy of both IWV and DA method is low. It is due to multiclass classification issue as we discussed earlier. From the comparison of different method it is evident that, lexical information plays a significant role in document sentiment classification. DA method incorporates domain information. However, it does not include lexical information. In our experiment DA does not make any significant score and outperformed by HIDE and IWV.

The cumilative analysis of comparision discussed above, gives an insight about each individual components of HIDE. For instance, dataset with larger vocabulary contain more domain specific information which helps HIDE to capture precise domain contexts. The domain information is also embedded into document level by LSA. The performance of HIDE in CR and SST dataset indicate that HIDE is relatively dependent on the nature of the training dataset. It also establishes the significance of document level domain information in HIDE. Lastly, lexical information corresponding to a word is as necessary as domain information to analyze the sentiment orientation of a document.

We further tallied between HIDE, IWV and DA to show the amount of improvement HIDE incur upon pre-trained GloVe embeddings. Table12 shows the comparision. Improvement using IWV is positive for MR, SST and CR datasets but negative for RT,SST-1 and ARD dataset. Whereas, HIDE shows consistent and positive increase over GloVe embedding for all the datasets.

**Table11: Comparison of HIDE with other existing methods**

| Dataset | Metrics | HIDE | IWV | DA |
|---|---|---|---|---|
| **MR** | Accuracy | **87.52** | 80.7 | 78.95 |
| | F-Score | **87.18** | 82 | 79.66 |
| **ARD** | Accuracy | **88.14** | 69 | 59.82 |
| | F-Score | **86.82** | 67.11 | 71.10 |
| **RT** | Accuracy | **83.00** | 82.3 | 54.21 |
| | F-Score | **83.50** | 81.0 | 68.00 |
| **CR** | Accuracy | 81.67 | **85.00** | 61.2 |
| | F-Score | 81.00 | **83.3** | 60 |
| **SST** | Accuracy | 81.74 | **87.0** | 57.50 |
| | F-Score | 79.32 | **84.8** | 62.53 |
| **SST-1** | Accuracy | **50.25** | 47.00 | 42.22 |
| | F-Score | **19.00** | 45.70 | 43.50 |

**Table12: Improvement over GloVe pre-trained embeddings based on accuracy (%)**

| Dataset | IWV – GloVe | DA- GloVe | HIDE-GloVe |
|---|---|---|---|
| **MR** | 1.7 | -0.05 | 8.52 |
| **ARD** | -9.42 | -18.6 | 9.72 |
| **RT** | -0.6 | -28.69 | 0.1 |
| **CR** | 7.5 | -17.5 | 4.17 |
| **SST** | 11.4 | -18.1 | 6.14 |
| **SST-1** | -3.00 | -7.78 | 0.25 |

## 6 Conclusion

Pre-trained word embeddings, which are used in various NLP tasks, suffers from poor accuracy in sentiment analysis tasks. Pre-trained word embeddings are trained on general purpose corpora which results in lack of domain and sentiment information in each word vectors. To address this problem, we propose an approach, Hybrid Improved Document level Embedding (HIDE). In HIDE, we improved word level embeddings with domain, sentiment and POS information. Each word of a document is combined into document level representation. Similarly, using LSA we generate document level embedding and combine both of these document level vectors for sentiment analysis. HIDE is tested using six benchmark datasets and it shows considerable improvement over pre-trained GloVe vectors and domain specific Word2Vec. We have tallied HIDE with two existing methods IWV and DA. HIDE comes as better solution for document level sentiment analysis in most of cases.

Although, experiments using HIDE gives good outcomes, it is not free from limitations. Firstly, we have used only GloVe and Word2Vec, there are few more well-known embeddings are available such as - Fasttext, Elmo etc. HIDE can be implemented using one of those embeddings as well. Secondly, the datasets involved in this experiment are not very large in size. Hence, experimentation can be done using large datasets. Thirdly, we conduct our experiment with movies and electronics domain, other domain datasets should also be tested. Moreover, deep learning techniques is not used in our experiments. With larger dataset and good deep learning models embeddings can be improved.